\documentclass{article} % For LaTeX2e
\usepackage{nips13submit_e,times}
\usepackage{hyperref}
\usepackage{url}
\usepackage{epsfig}
\usepackage{blindtext}

\pdfoutput=1

\title{ Occupancy Detection in Vehicles Using Fisher Vector Image Representation}

\author{
Yusuf Artan\\
Xerox Research Center\\
Webster, NY 14580 \\
\texttt{Yusuf.Artan@xerox.com} \\
\AND
Peter Paul\\
Xerox Research Center\\
Webster, NY 14580 \\
\texttt{Peter.Paul@xerox.com} \\
}

\nipsfinalcopy % Uncomment for camera-ready version

\begin{document}

\maketitle

\begin{abstract}
Due to the high volume of traffic on modern roadways, transportation agencies have proposed High Occupancy Vehicle (HOV) lanes and High Occupancy Tolling (HOT) lanes to promote car pooling. However, enforcement of the rules of these lanes is currently performed by roadside enforcement officers using visual observation. Manual roadside enforcement is known to be inefficient, costly, potentially dangerous, and ultimately ineffective. Violation rates up to $50\%-80\%$ have been reported, while manual enforcement rates of less than $10\%$ are typical. Therefore, there is a need for automated vehicle occupancy detection to support HOV/HOT lane enforcement. A key component of determining vehicle occupancy is to determine whether or not the vehicle's front passenger seat is occupied. In this paper, we examine two methods of determining vehicle front seat occupancy using a near infrared (NIR) camera system pointed at the vehicle's front windshield. The first method examines a state-of-the-art deformable part model (DPM) based face detection system that is robust to facial pose. The second method examines state-of-the-art local aggregation based image classification using bag-of-visual-words (BOW) and Fisher vectors (FV). A dataset of 3000 images was collected on a public roadway and is used to perform the comparison. From these experiments it is clear that the image classification approach is superior for this problem. 
\end{abstract}

\section{Introduction}
Intelligent transportation systems (ITS) improve safety and mobility through the integration of sensing, computational power, and advanced communications into the transportation infrastructure \cite{Xerox}. Managed lanes, enabled by ITS, combine traffic management, tolling, transit, and carpooling in a multi-purpose roadway, creating new options for agencies and the traveling public, and use congestion pricing to more efficiently manage demand and generate revenue. 
High Occupancy Vehicle (HOV) lanes are standard car-pool lanes where either two (2+) or three (3+) vehicle occupants are required in order to use the lane. Due to these requirements, the HOV lanes are typically less congested and allow a vehicle to make its journey more rapidly \cite{Georgia}. High Occupancy Tolling (HOT) lanes are a form of managed lanes where single occupant vehicles are allowed to use the HOV lane upon payment of a toll. Typically the toll is dynamically adjusted in real-time to maintain a minimum speed on the roadway - often the minimum speed is 45 miles per hour. If the average traffic speed on the HOT lane starts to become lower than 45 mph, the toll prices increases to discourage additional vehicles from entering the HOT lane. However, to realize the congestion reducing benefit of HOV/HOT lanes, the rules of the HOV/HOT lane need to be enforced. 
To enforce the rules of these lanes, current practice requires dispatching law enforcement officers at the roadside to visually examine incoming cars. This method is expensive, difficult, and ultimately ineffective. Typical violation rates can exceed 50\%-80\%, while manual enforcement rates are typically less than 10\% \cite{Stephen}. Currently, transportation authorities are seeking an automatic or semi-automatic approach to replace and/or augment this manual enforcement. Any automatic system would require very high accuracies. A typical automatic license plate reading system, for example, has over 99.9\% accuracy while making a decision on over 80\% of the vehicles.
Image based vehicle occupancy detection systems have been examined by past researchers \cite{Birch}, \cite{Hao}, \cite{Georgia}, \cite{Billheimer}, \cite{Peter1}. For automated occupancy detection, face detection was used in these earlier studies. More recently, studies have proposed methods that extract certain features to count passengers including faces, seats, seat belts, or skin that are visible to the camera to distinguish driver + passenger vs. driver only \cite{Peter2}.

\begin{figure*}[t!]
\centerline{
\begin{tabular}{c@{}c@{}}
\includegraphics[width=2.64in, height=2.4in]{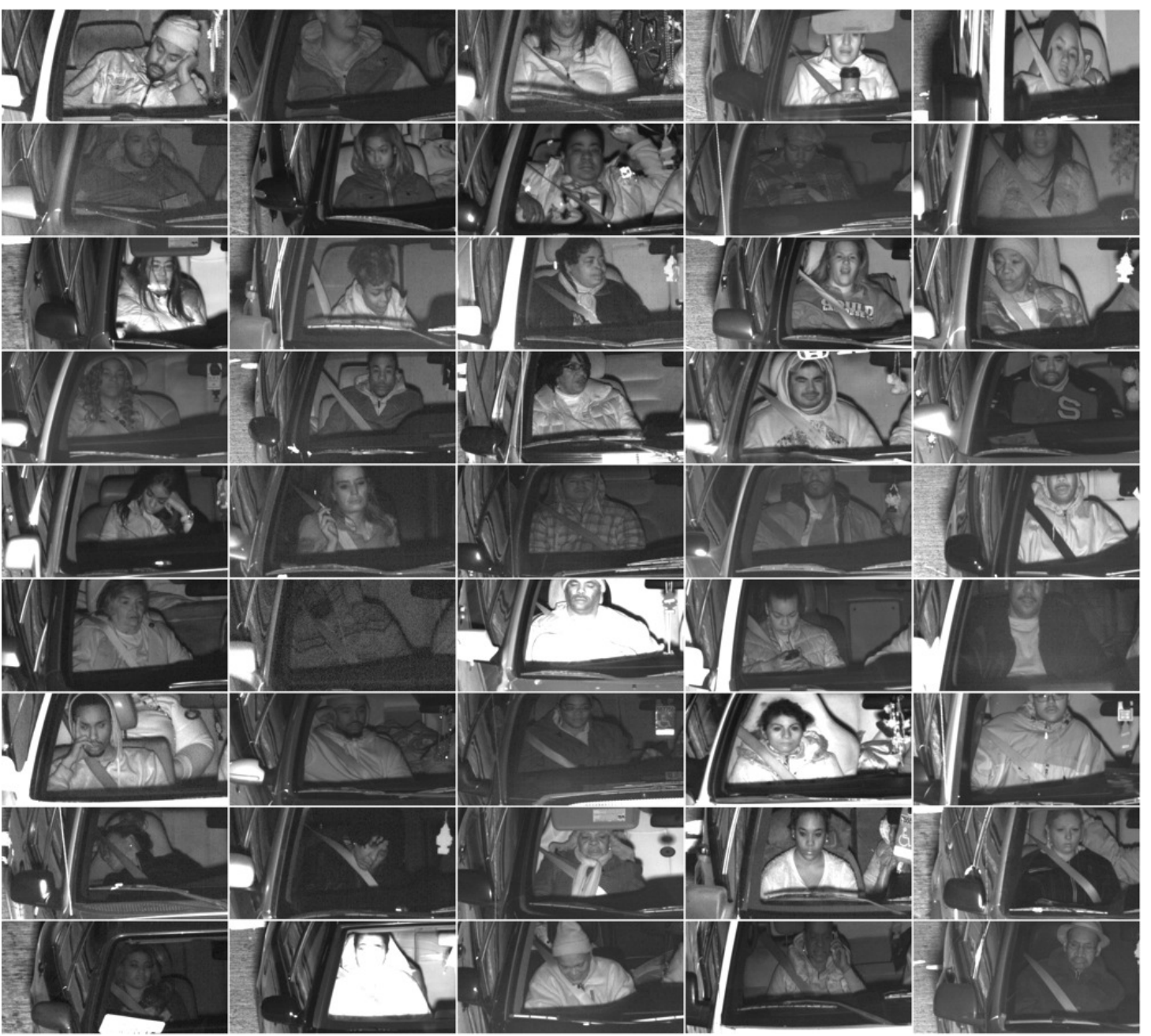} &\includegraphics[width=2.64in, height=2.4in]{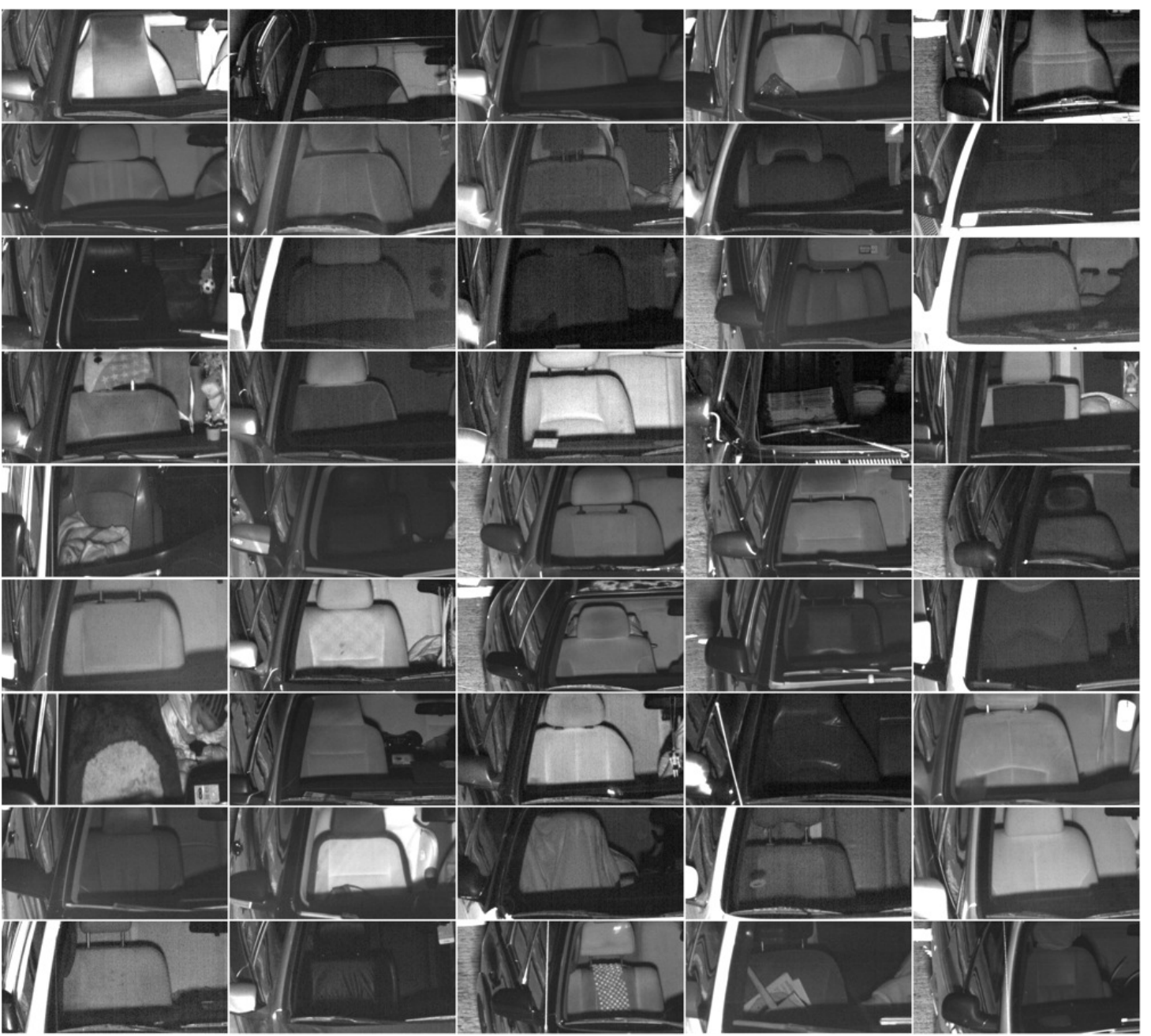} \\
(A) & (B) \\
\end{tabular} }
\caption{ \textbf{Overview of the occupancy image database. Part (A) is the passenger side images captured using NIR camera
system. Part (B) shows the empty seat images.} }
\label{fig:1}
\end{figure*}
In this paper, we examine front seat vehicle occupancy detection for use as part of an automatic or semi-automatic HOV/HOT lane enforcement system. Other parts of the HOV/HOT lane enforcement system would include window/windshield locators and rear seat detection. However, in this work we focus on front passenger detection. An empty front passenger seat may indicate a 'candidate violator' such that the imagery is further reviewed and processed to determine its final violator status. Our experimental setup includes a camera based imaging system to capture still images in the near-infrared (NIR) band through the windshield of an approaching vehicle. NIR imaging is commonly used in transportation imaging systems so that the illumination does not distract drivers.\newline
\indent
Two detection approaches are examined in this work. As in past studies, the first approach uses face detection methods to perform front seat occupancy detection. The second approach uses a machine-learning based classifier that detects front seat occupancy by using a global image representation of the captured images.
In \cite{Zhu}, authors developed a model that is a novel and simple approach to encode the elastic deformation and 3D structure of an object for face detection and pose estimation. It uses mixtures of trees (poses) with a shared pool of parts defined at each landmark positions. It then uses global mixtures to model topological changes due to view points.  That study showed state-of-the-art results for faces larger than $80\times80$ pixels.
In this paper, we use this deformable part models to detect front seat passenger faces and compare results to image classification based methods.
\newline
\indent
A popular image classification method is to describe images with bag-of-visual-words (BoW) histograms and to classify them using non-linear support vector machines (SVMs) \cite{Csurka}. In the BoW representation, dense local descriptors (such as SIFT \cite{SIFT}) are extracted from the image and each descriptor is assigned to its closest visual word in a visual codebook obtained by clustering a large set of descriptors with k-means \cite{Sivic}, \cite{Jegou}. In recent studies, Fisher vectors (FV) have shown to yield better performance in image categorization tasks than BoW \cite{Perronin1}, \cite{Perronin3}. Therefore, we utilize FV image representation to perform image classification in front seat occupancy detection task.\newline
\indent
The organization of this article as follows. In Section \ref{sec:Methodology}, we briefly describe the details of the deformable part based face detection method and FV descriptors for image classification. Evaluation of the methods using real world road images are presented in Section \ref{sec:Experiments}. In Section \ref{sec:Discussion}, we discuss the advantages \& disadvantages of the proposed approach and present the conclusions. 
\section{Image DataSet:}
\textbf{Dataset:} We have collected a large roadway dataset using an NIR imaging system. The distance from camera to cars is on average 60-feet and horizontal field of view of the camera is approximately 12-feet at that distance. In these experiments, we use 3000 vehicle front passenger seat images along with their ground truth.
\newline
\indent
Figure \ref{fig:1} illustrates a set of passenger side NIR images with and without passengers from the dataset. Note that there is a large variation in image intensity due to variations in windshield transmission, illuminator state, sun position, cloud position, and other factors. The passenger images show a wide variety of human faces, facial poses, occlusions, and other expected 'within class' variations. The empty seat images also show a large 'within class' variation. Face detection is a natural approach for this problem since the vehicle occupants have visible faces. Past researchers have taken this approach. However, we observe, that due to the large variations within the image dataset of real-world vehicle images, face detection is challenged. Further, the front passenger seat detection problem is inherently a binary classification problem where either 1 or 0 faces exist in the scene. Thus an image classification approach is indicated.

\section{Methodology} \label{sec:Methodology}
In this section, we first briefly describe the face detection method of \cite{Zhu}. Next, we present the FV representation, which aggregate the $d$-dimensional local descriptors into a single vector representation.
\subsection{Deformable Part Based Face Detection}
In a recent study, \cite{Zhu} developed a face detection method based on a mixture of trees with a set of parts $V$. Every facial part is considered as a node (part) in the tree and a mixture model is used to capture topological changes due to viewpoint. In the tree-structured model, each tree $T^{m}=(V^{m},E^{m})$ is linearly parameterized pictorial model, where $m$ indicates the mixture, $V^{m}$ is the set of parts ($V^{m}\subseteq V$) and $E^{m}$ is the set of edges between parts. \cite{Zhu} defined a score for a particular configuration of parts $L=\{l_{i}; i\in V\}$ for the given image $I$ as shown in Eq. \ref{eq:DPM1}, 
\begin{eqnarray} \label{eq:DPM1}
S(I,L,m) &=& App_{m}(I,L) + Shape_{m}(L) \nonumber \\
&=& \sum_{i \in V^{m}}w^{m}_{i} \cdot \phi(I,l_{i}) \nonumber \\
&+& \sum_{ij \in E^{m}} a^{m}_{ij}dx^2+b^{m}_{ij}dy^2+c^{m}_{ij}dx+d^{m}_{ij}dy \nonumber \\
\end{eqnarray}
where $\phi{(I,l_{i})}$ is the histogram of gradients (HoG) features (for the landmark points)  extracted at pixel location $l_{i}$ \cite{Dalal}, and $l_{i}=(x_{i},y_{i})$ for the pixel location of part $i$. $App$ term sums the appearance evidence for placing the $i^{th}$ template, $w^{m}_{i}$, tuned for mixture $m$, at location $l_{i}$. $Shape$ term score the spatial arrangement of the set of parts $L$, where $dx$ ($dy$) term represents the spatial deformation in $x$ ($y$) axis between parts $i$ and $j$. This model can be viewed as a linear classifier \cite{Felzenszwalb} with unknown parameters $w^{m}_{i}$ and $\{a^{m}_{ij},b^{m}_{ij},c^{m}_{ij},d^{m}_{ij}\}$ learned during training using latent SVM as shown in \cite{Felzenszwalb}.  During inference, we maximize Eq. \ref{eq:DPM2} over $L$ and $m$ for a given test image $I$ using dynamic programming to find the best configuration of parts as shown in \cite{Zhu}, \cite{Felzenszwalb}.  
\begin{equation} \label{eq:DPM2}
S^{*}(I) = \max_{m} [ \max_{L} S(I, L,m) ]
\end{equation}

\subsection{Image Classification Based Person Detection}
In image classification, statistics generated from local invariant descriptors are aggregated into  an image level vector signature, which is subsequently passed into a classifier. FV has recently become a standard for the aggregation part \cite{Perronin3}.\newline
\indent 
FV has been proposed to incorporate generative models into discriminative classifiers \cite{Jaakkola}.  Suppose $X=\{x_{t}, t=1...T\}$ denote the set of $T$ local descriptors extracted from a given image. We assume that generation process of local descriptors can be modeled by a probabilistic model $p(X|\theta)$ where $\theta$ denotes the parameters of the function. \cite{Jaakkola} proposed to describe $X$ by the gradient vector :
\begin{equation}
G_{\theta}^{X} = \frac{1}{T} \nabla_{\theta} \log p(X|\theta)
\label{eq:fishervector}
\end{equation}
, in which the gradient of the log-likelihood describes the contribution of the parameter $\theta$ to the generation process. Its dimensionality only depends on the number of parameters in $\theta$.\newline
\indent 
A natural kernel on these gradient vectors is the fisher kernel \cite{Jaakkola},
\begin{equation}
K(X,Y) = G_{\theta}^{X^{T}}F_{\theta}^{-1}G_{\theta}^{Y}
\end{equation}
where $F_{\theta}$ is the Fisher information matrix of $p(X|\theta)$:
\begin{equation}
F_{\theta} = E_{x \sim p}[ \nabla_{\theta}\log p(X|\theta) \nabla_{\theta}\log p(X|\theta)^{T}]
\end{equation}
where $F_{\theta}^{-1}$ is symmetric and positive definite, it has a Cholesky decomposition $F_{\theta}^{-1}=L_{\theta}^{T}L_{\theta}$. Therefore the kernel $K(X,Y)$ can be written as a dot product between normalized vectors ($\textit{g}_{\theta}$) shown in Eq.  \ref{eq:FV1}.
 \begin{equation}\label{eq:FV1}
 \textit{g}^{X}_{\theta} = L_{\theta}G_{\theta}^{X}
\end{equation}
Typically, $\textit{g}^{X}_{\theta}$ is referred to as fisher vector of $X$.
Similar to the earlier work \cite{Perronin1}, we assume that $p(X|\theta)$ is a gaussian mixture model (GMM): $p(x|\theta)=\sum_{i=1}^{K}w_{i}p_{i}(x)$. We denote $\theta = \{w_{i},\mu_{i},\sigma_{i}, i=1...K \}$ where $w_{i},\mu_{i}$, and $\sigma_{i}$ are respectively the mixture weight, mean vector and variance matrix (assumed diagonal) of Gaussian $p_{i}$.
In this paper, we only consider gradients with respect to the mean. We use the diagonal closed form approximation of the Fisher information matrix of \cite{Perronin1} in which case the normalization of the gradient by $L_{\theta}$ is simply a whitening of the dimensions.\newline
\indent
Let $\alpha_{i}(t)$ be the assignment of the descriptor $x_{t}$ to the $i^{th}$ gaussian:
\begin{equation}
\alpha_{i}(t) = \frac{w_{i}p_{i}(x_{t}|\theta)}{\sum_{j=1}^{K}w_{j}p_{j}(x_{t}|\theta)}
\end{equation}
Let $\textit{g}_{i}^{X}$ denotes the $d$-dimensional gradient with respect to the mean $\mu_{i}$ of Gaussian $i$.  
Assuming that the $x_{t}$\rq{}s are generated independently by $p(X|\theta)$, we obtain Eq. \ref{eq:8} 
after mathematical derivations:
\begin{equation}\label{eq:8}
\textit{g}_{i}^{X} = \frac{1}{T\sqrt{w_{i}} }\sum_{t=1}^{T}\alpha_{t}(i) ( \frac{x_{t}-\mu_{i}}{\sigma_{i}} )
\end{equation}
The final vector $\textit{g}_{\theta}^{X}$ is the concatenation of the $\textit{g}_{i}^{X}$ vectors for $i=\{1..K\}$ and is $K\times d$ dimensional. Experiments have been performed for values ranging from $K=32$ to $K= 512$.\newline
\indent
As noted in earlier studies, bag of words required $d$ times less visual words, however, our experiments indicate that FK based learning yields better performance as shown in Section \ref{sec:Experiments}. In this paper, we utilize stochastic gradient descent (SGD) SVM classifier to construct the classification model \cite{Zeynep}, \cite{Bottou}.

\section{Experiments} \label{sec:Experiments}
\subsection{Face Detection Performance}
\begin{figure}[!t]
 \centerline{
\begin{tabular}{c@{}c@{}c@{}c@{}c@{}}
\includegraphics[width=1in, height=1in]{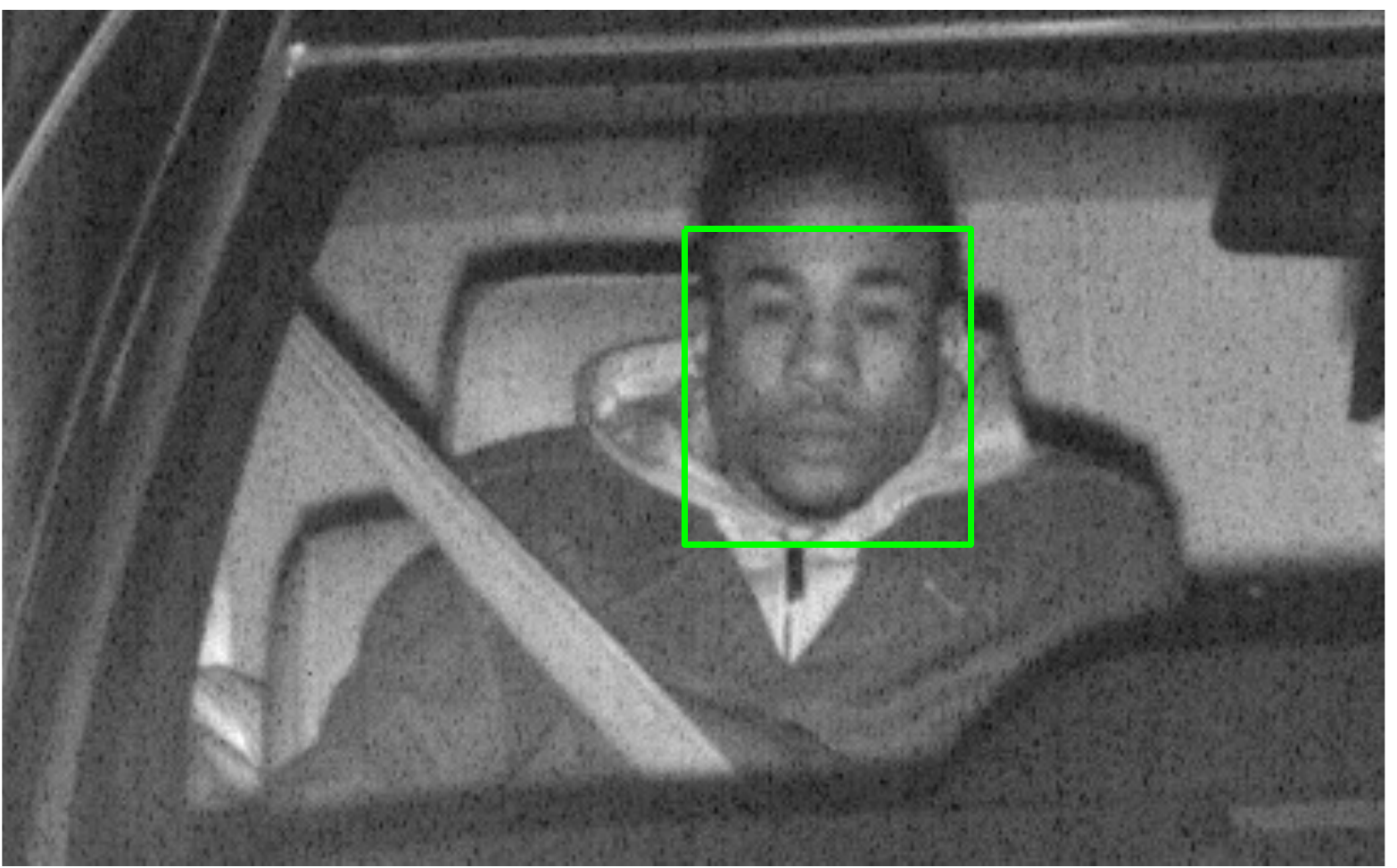} & \includegraphics[width=1in, height=1in]{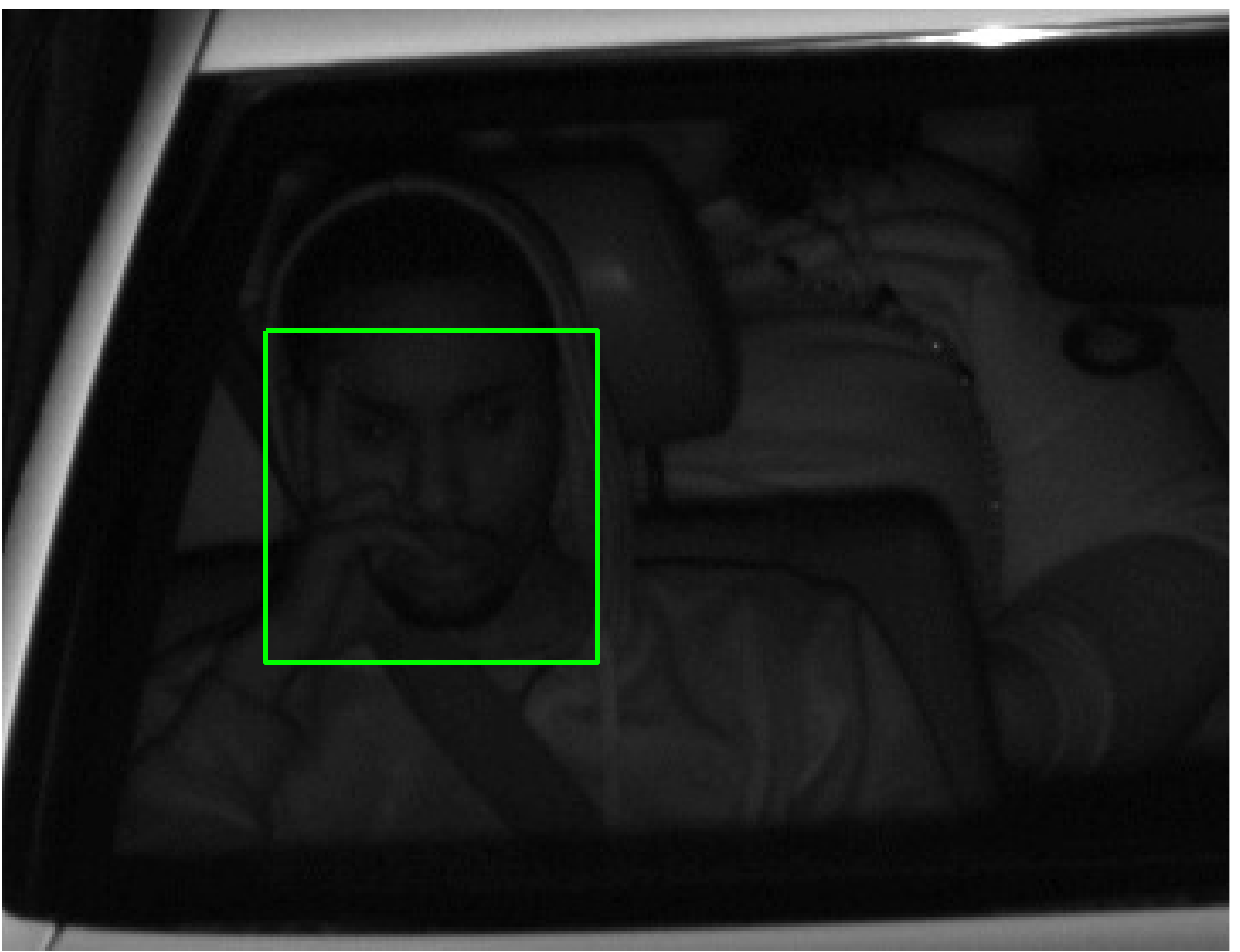} & \includegraphics[width=1in, height=1in]{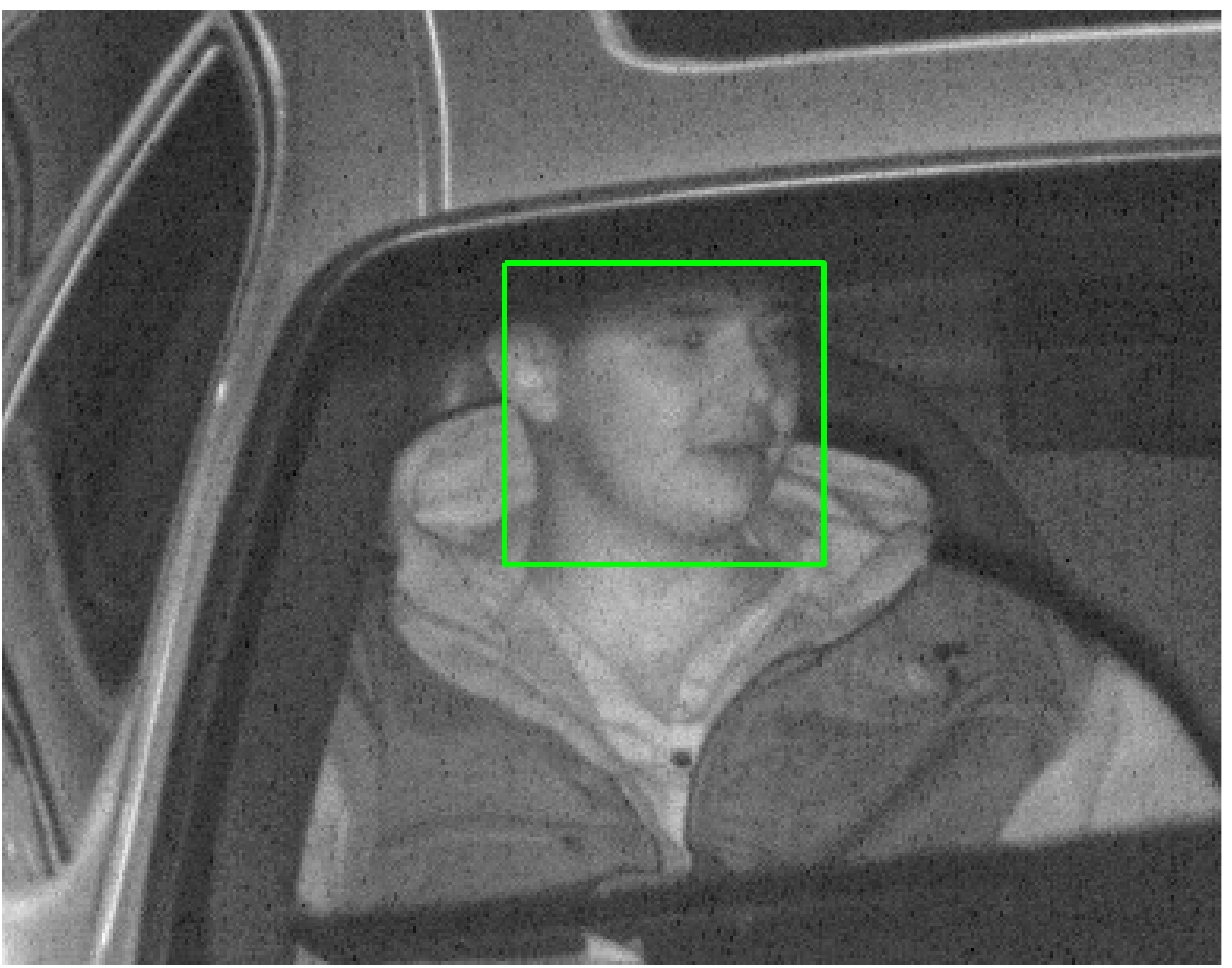} &
 \includegraphics[width=1in, height=1in]{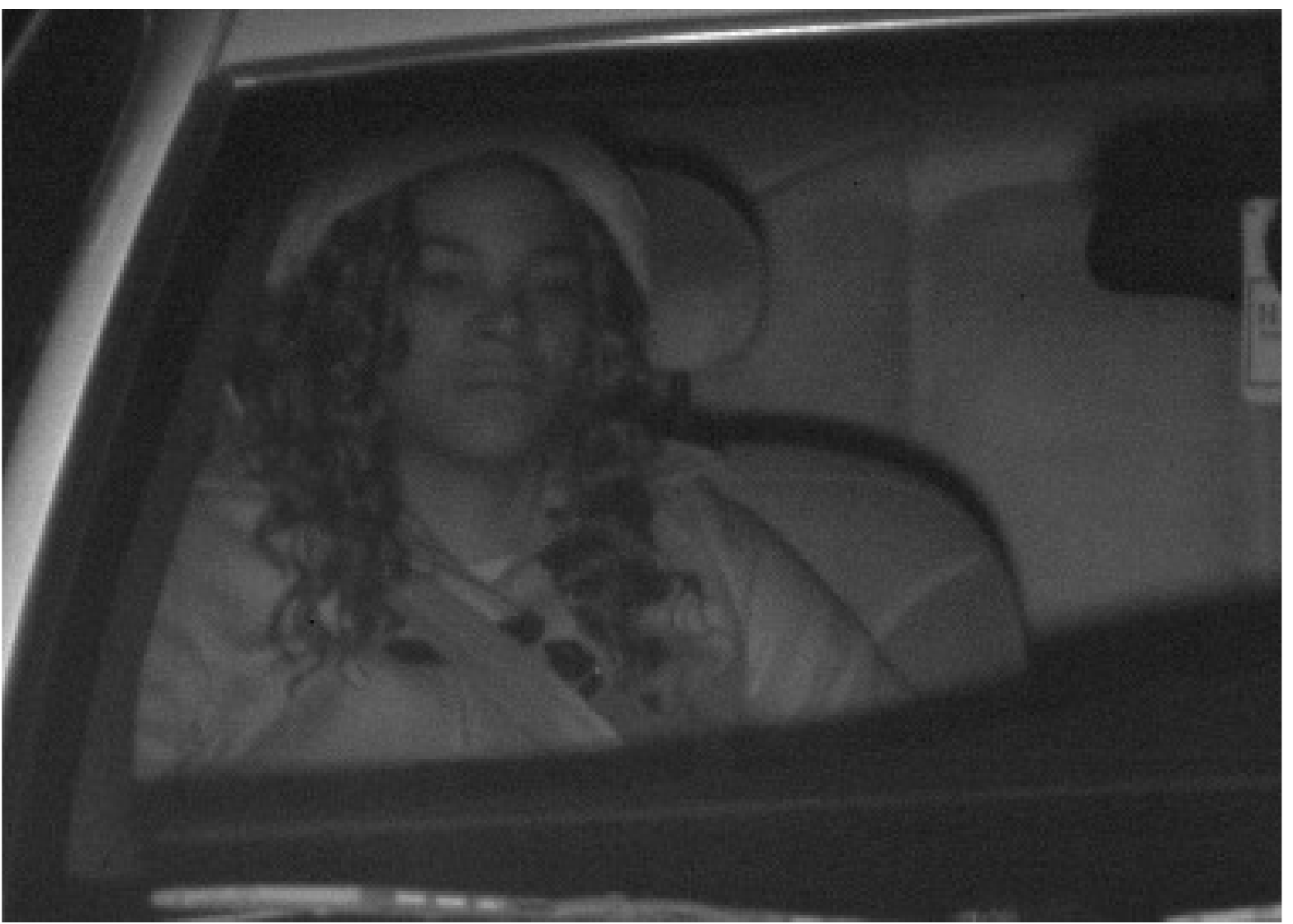} & \includegraphics[width=1in, height=1in]{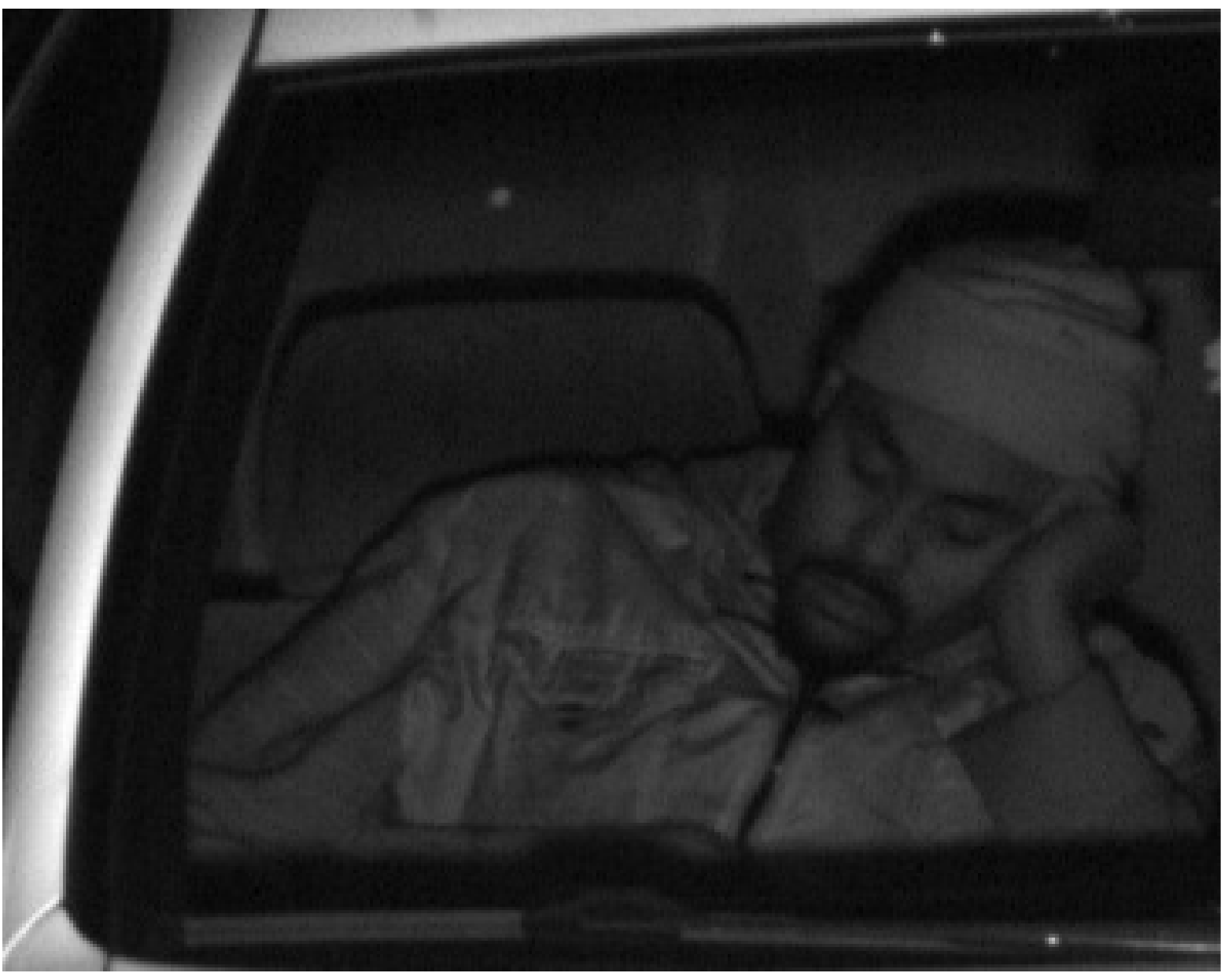} \\
\end{tabular}
}
\caption{ \textbf{State-of-the-art face detector \cite{Zhu} applied to some test images in our dataset. Face detector misses the faces shown in the last 2 columns.}  }
\label{fig:FaceDetection}
\end{figure}

Before presenting the experimental results using image classification techniques, let us first present results using state-of-the-art face detector in our application \cite{Zhu}. Face detection method \cite{Zhu} has shown promising performance for faces with arbitrary poses superior to the classical Viola-Jones face detector \cite{Jones}. However, occlusion, illumination changes and camera conditions results in high miss rate using this technique.
In this study, we define an overlap measure ($\mathrm{overlap} = \mathrm{area}|A\cap B|/\mathrm{area}|A\cup B|$, where $A$ and $B$ are the segmentation output and ground truth, respectively) to quantify an error rate by comparing a candidate face area to the ground truth. If the overlap is larger than 0.6, we declare the (face) detection as true positive.\newline
\indent
For instance, Figure \ref{fig:FaceDetection} presents the sample image results obtained using landmark based deformable part models for face detection. Results indicate that face detection works well when details of face region is not occluded, however, it fails in arbitrary view angles and under occlusion. For our test set, we achieve an accuracy of 92.3\% using the face detection technique of \cite{Zhu}. This is the baseline approach to compare our image classification based results.
\subsection{Occupancy Detection:}
In this section, we present a comparison of occupancy detection using various local aggregation methods (FV, BoW, and VLAD) as well as state-of-the-art face detection techniques. For local aggregation based methods, we extract features from $24\times24$ pixel patches on regular grids ( every 4 pixels) at 3 scales.
We only extract 128-D SIFT descriptors for these image patches. Next, we utilize principal components analysis (PCA) to reduce SIFT dimensions to 64.   
\subsubsection{Fisher Vectors (FV):} 
In our experiments, we used gaussian mixture models (GMM) with $K = \{32, 64, 128, 256, 512\}$ gaussians to compute the FVs. The GMM's are trained using the maximum likelihood (ML) criterion and a standard expectation maximization (EM) algorithm. Similar to \cite{Perronin2}, we apply the power and L2 normalization to fisher vectors to improve the classification performance.
\subsubsection{Bag-of-Words (BoW):}
The BoW representation is a ubiquitously method to group together descriptors \cite{Sivic}, \cite{Csurka}. It requires the definition of a codebook of $K$-centroids (visual words) obtained by k-means clustering.  Each local descriptor of dimension $d$ from an image is assigned to the closest centroid. BoW is the histogram of the count of visual words assigned to each visual word. Therefore it produces a K-dimensional vector, which is subsequently normalized. In this study, we follow a spatial pyramid based BoW representation in which we create histograms for the original image and spatial grid partitioning of the original image. In our studies, we concatenate histograms from the original image, $2\times2$ and $4\times4$ regular spatial grid. 
\subsubsection{Vector of Locally Aggregated Descriptors (VLAD):}
\begin{figure}[!t]
 \centerline{
\begin{tabular}{c@{}}
 \includegraphics[width=2.6in, height=2.2in]{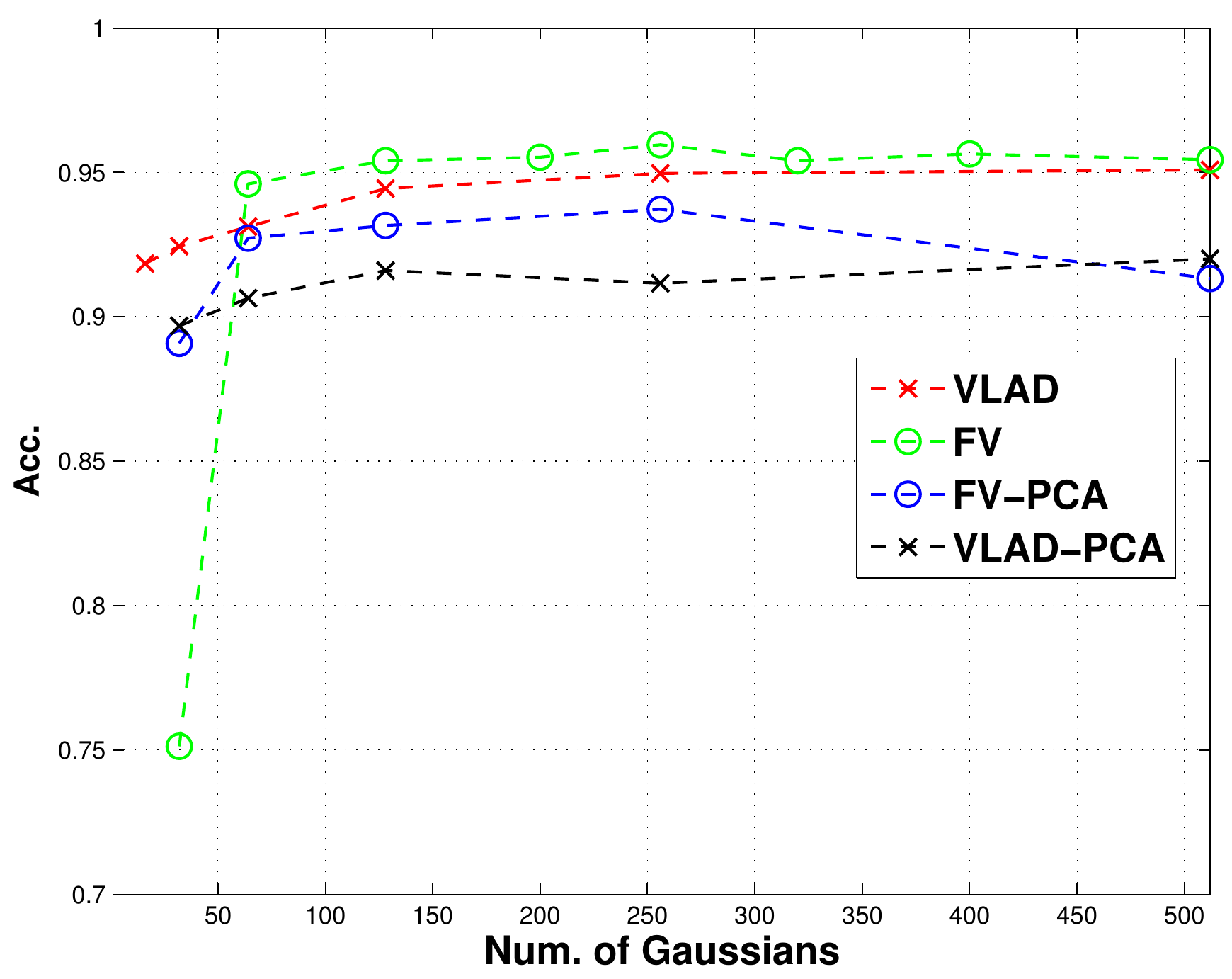} \\
\end{tabular}
}
\caption{ \textbf{Comparison of FV and VLAD representations for various number of gaussian components as well as the PCA applied FV and VLAD representations.} }
\label{fig:VLADvsFV}
\end{figure}

\begin{figure}[!t]
\centering{
\begin{tabular}{c@{}c@{}}
\includegraphics[width=2.6in, height=2.2in]{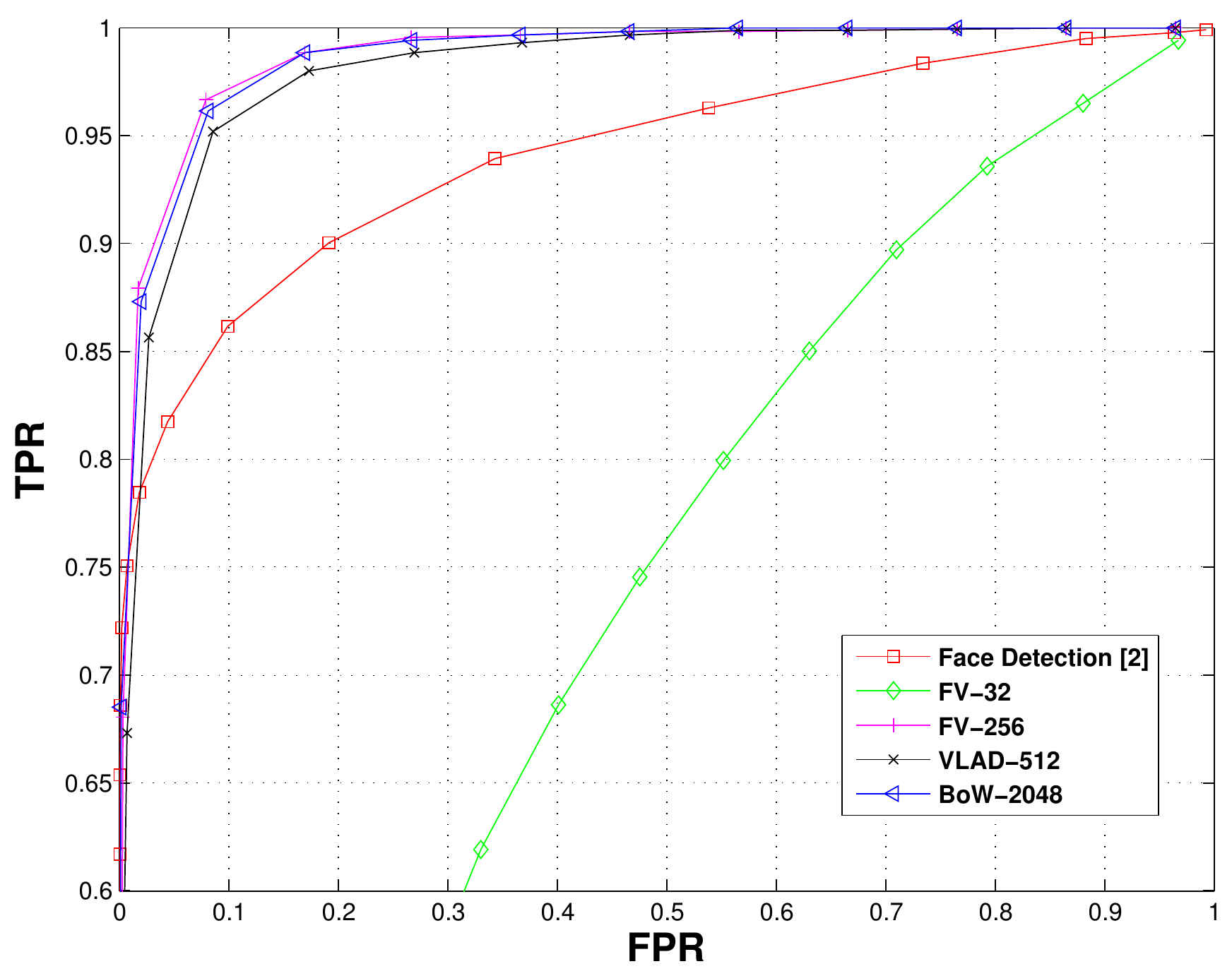} & \includegraphics[width=2.6in, height=2.2in]{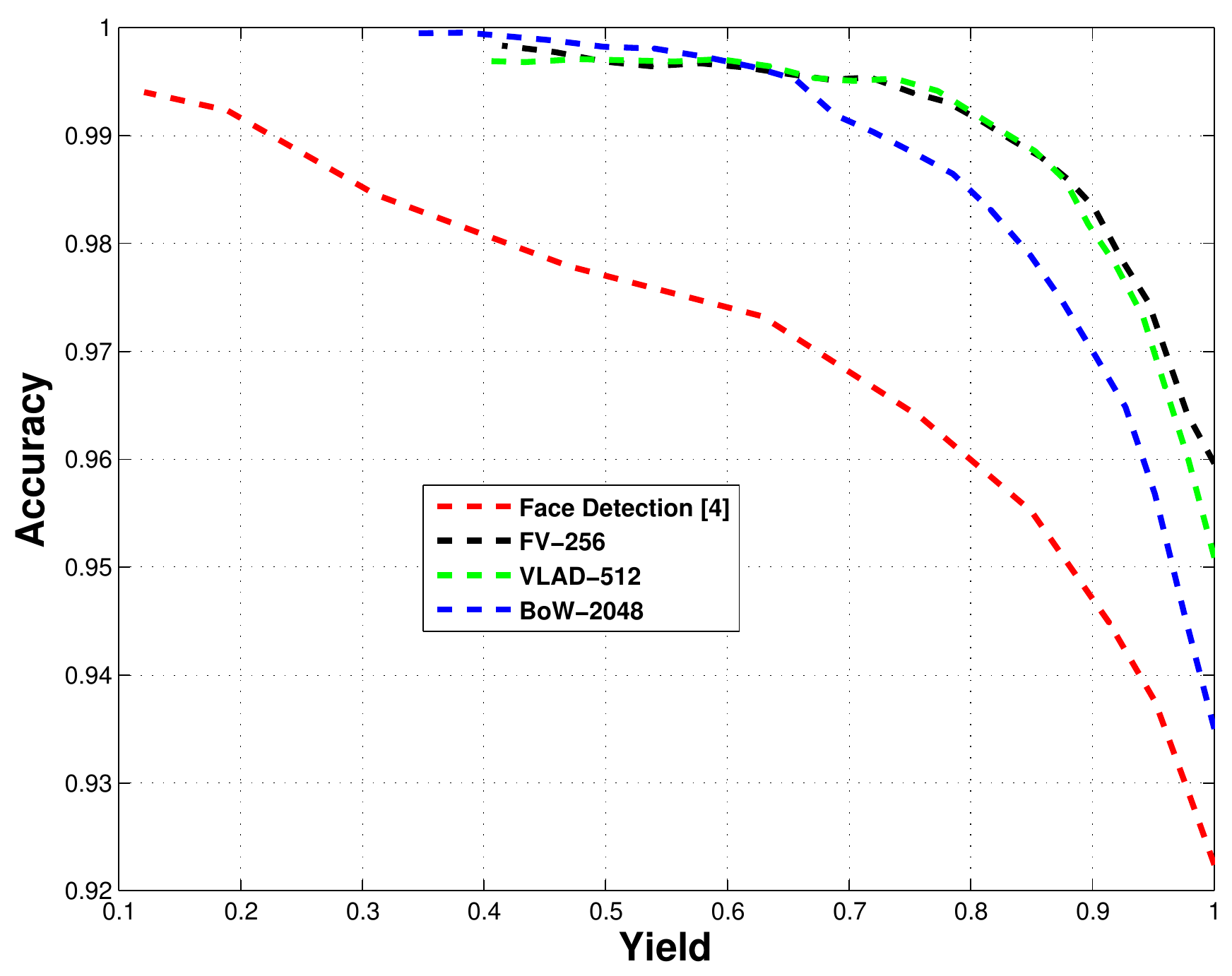}\\
 (A) & (B) \\
 \end{tabular}
}
\caption{ \textbf{Part (A) presents comparison of ROC curves for SVM classifiers constructed using FV, BoW, VLAD features and face detection method of \cite{Zhu}. Part (B) show the accuracy versus yield for SVM classifiers constructed using FV, BoW, VLAD features and face detection \cite{Zhu}. } }
\label{fig:ROCs}
\end{figure}

\begin{figure*}[!t]
\centering{
\begin{tabular}{|c|c|c|}
\hline
Image & Face Detect. \cite{Zhu} &  FV \\
\hline
\includegraphics[width=0.8in, height=0.6in]{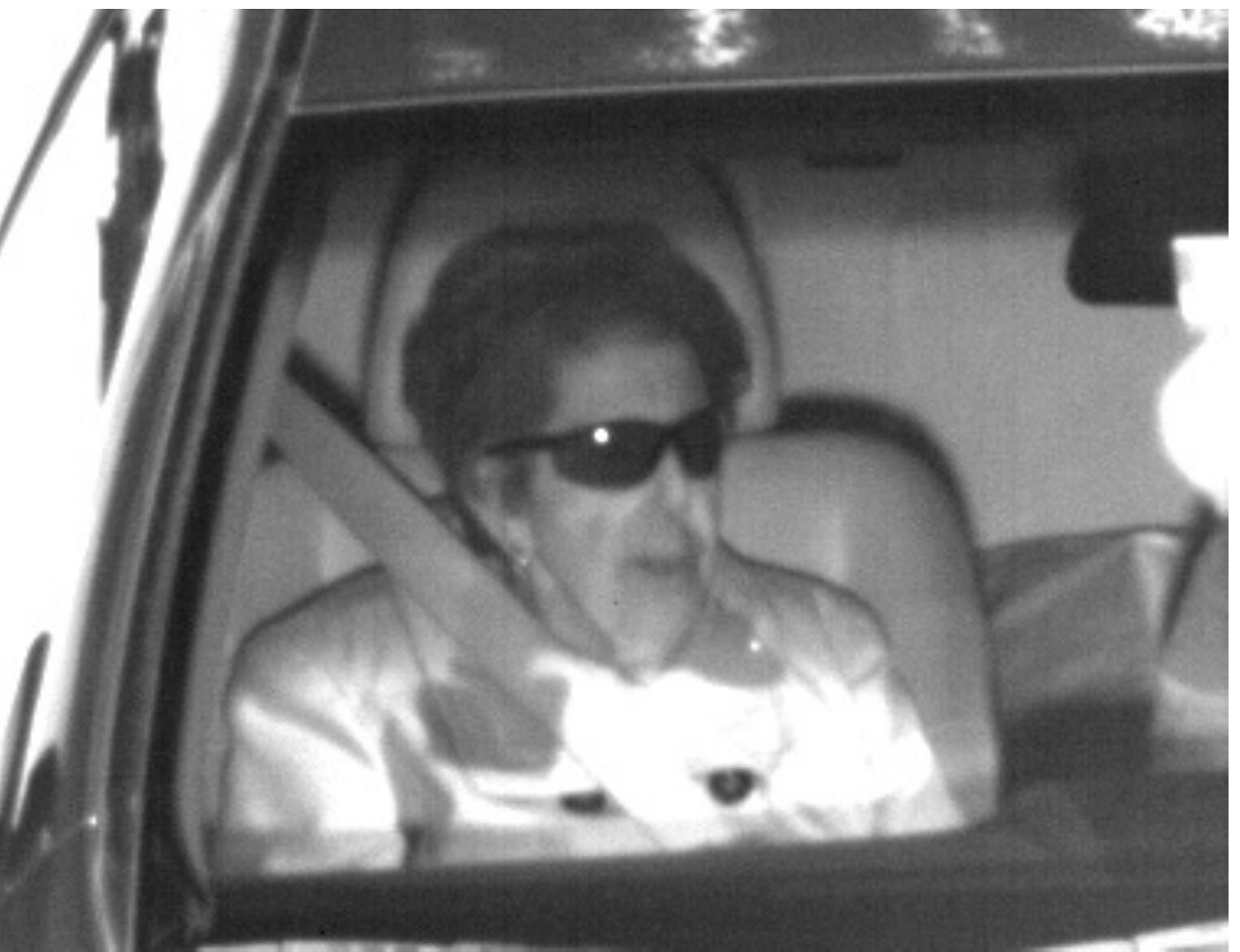} &  Empty & Person  \\ 
\hline
\includegraphics[width=0.8in, height=0.6in]{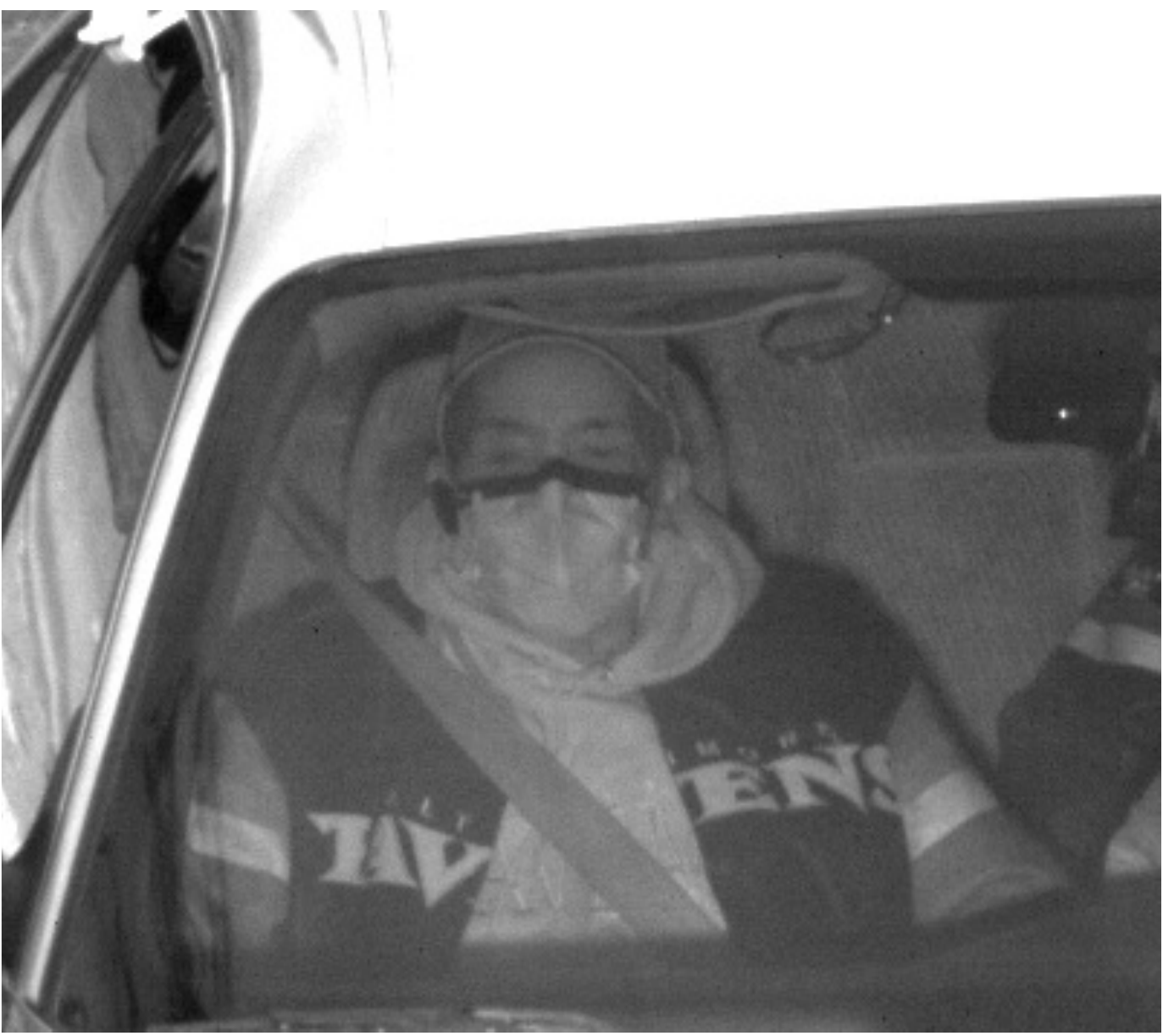} &  Empty & Person\\ 
\hline
\includegraphics[width=0.8in, height=0.6in]{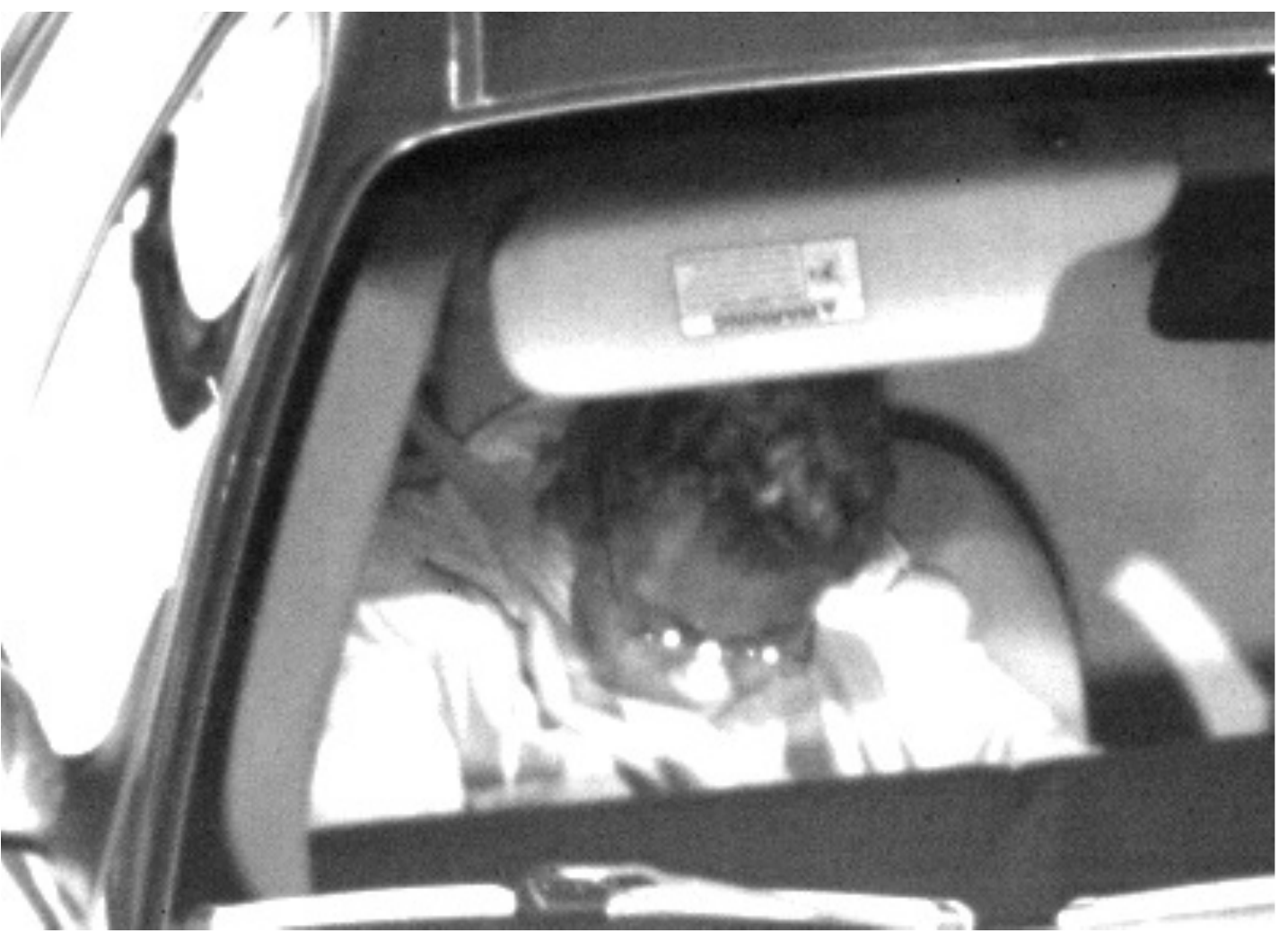} & Empty & Person \\ 
\hline
\includegraphics[width=0.8in, height=0.6in]{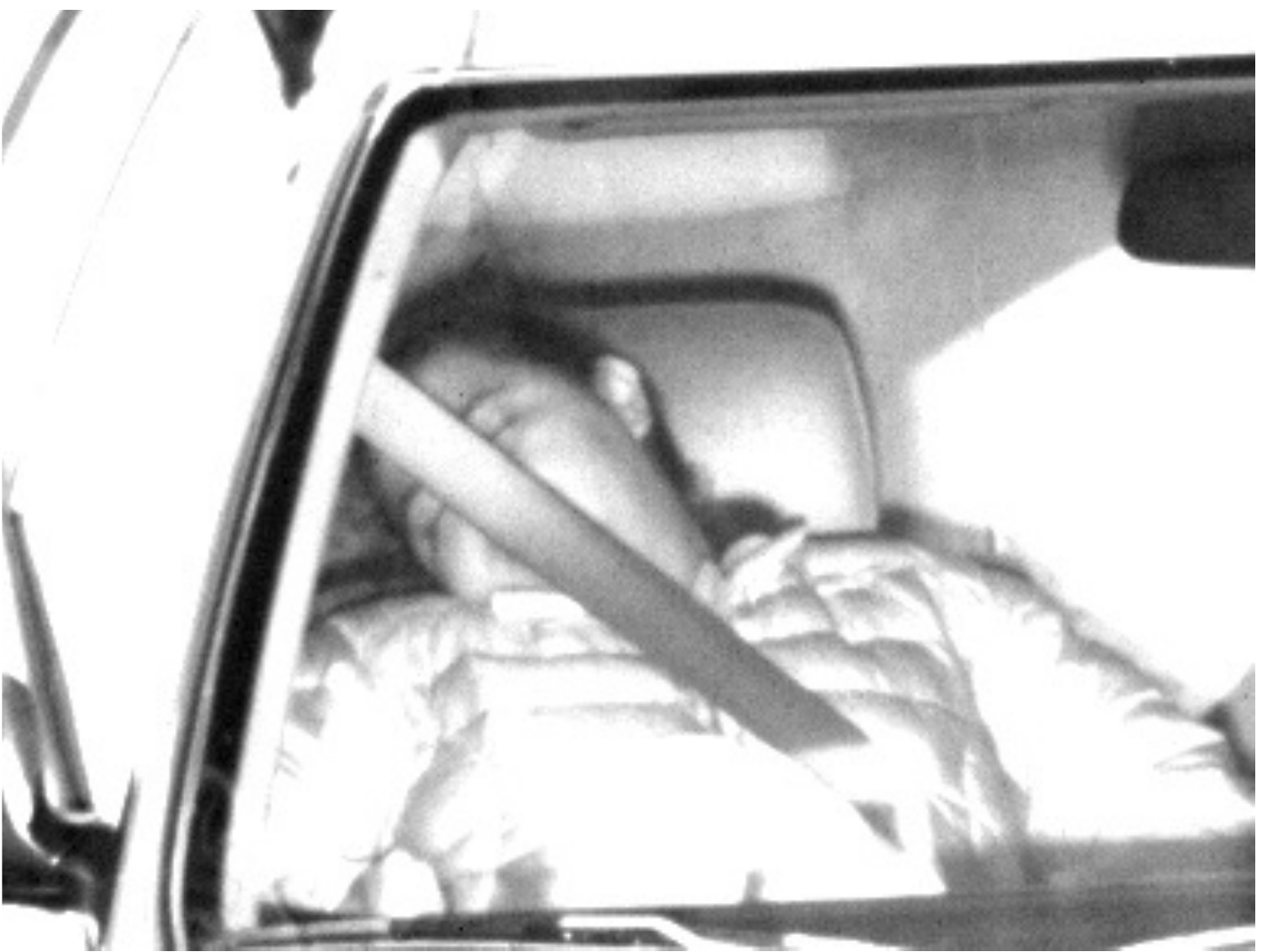} &  Empty & Person \\ 
\hline
\includegraphics[width=0.8in, height=0.6in]{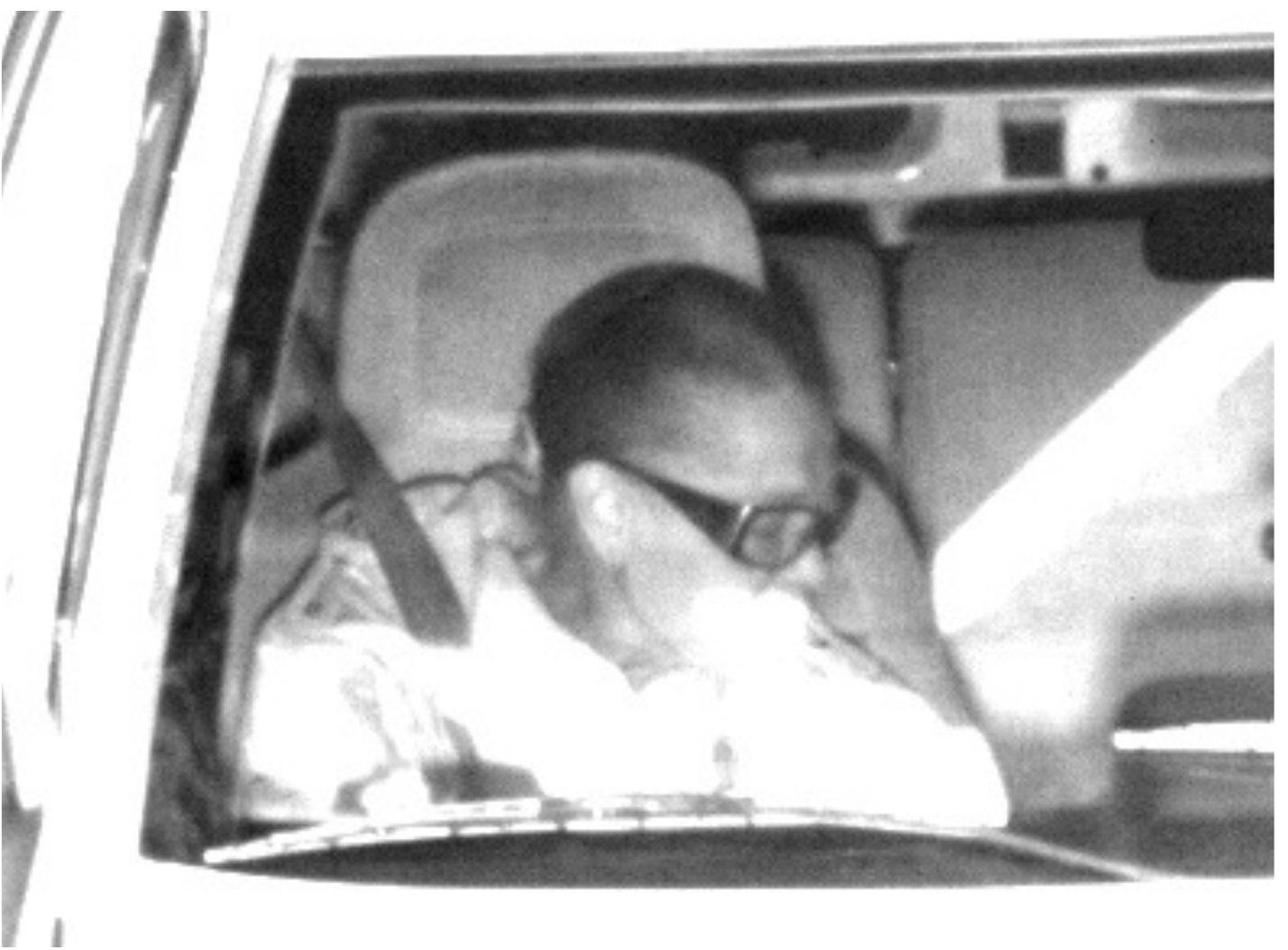} & Empty & Person  \\ 
\hline
\end{tabular}
\begin{tabular}{|c|c|c|}
\hline
Image & Face Detect. \cite{Zhu} & FV \\
\hline
\includegraphics[width=0.8in, height=0.6in]{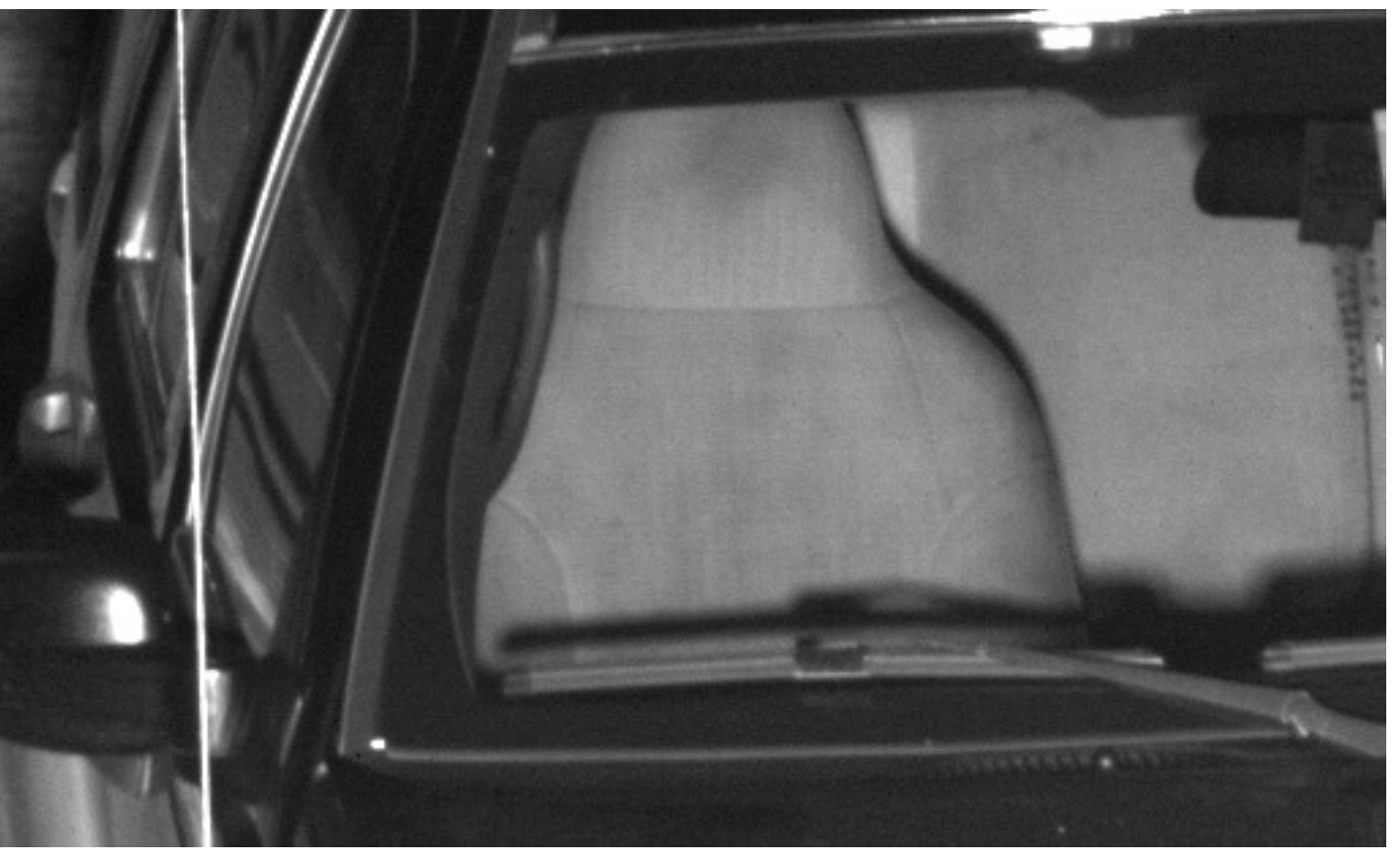} & Person & Empty \\ 
\hline
\includegraphics[width=0.8in, height=0.6in]{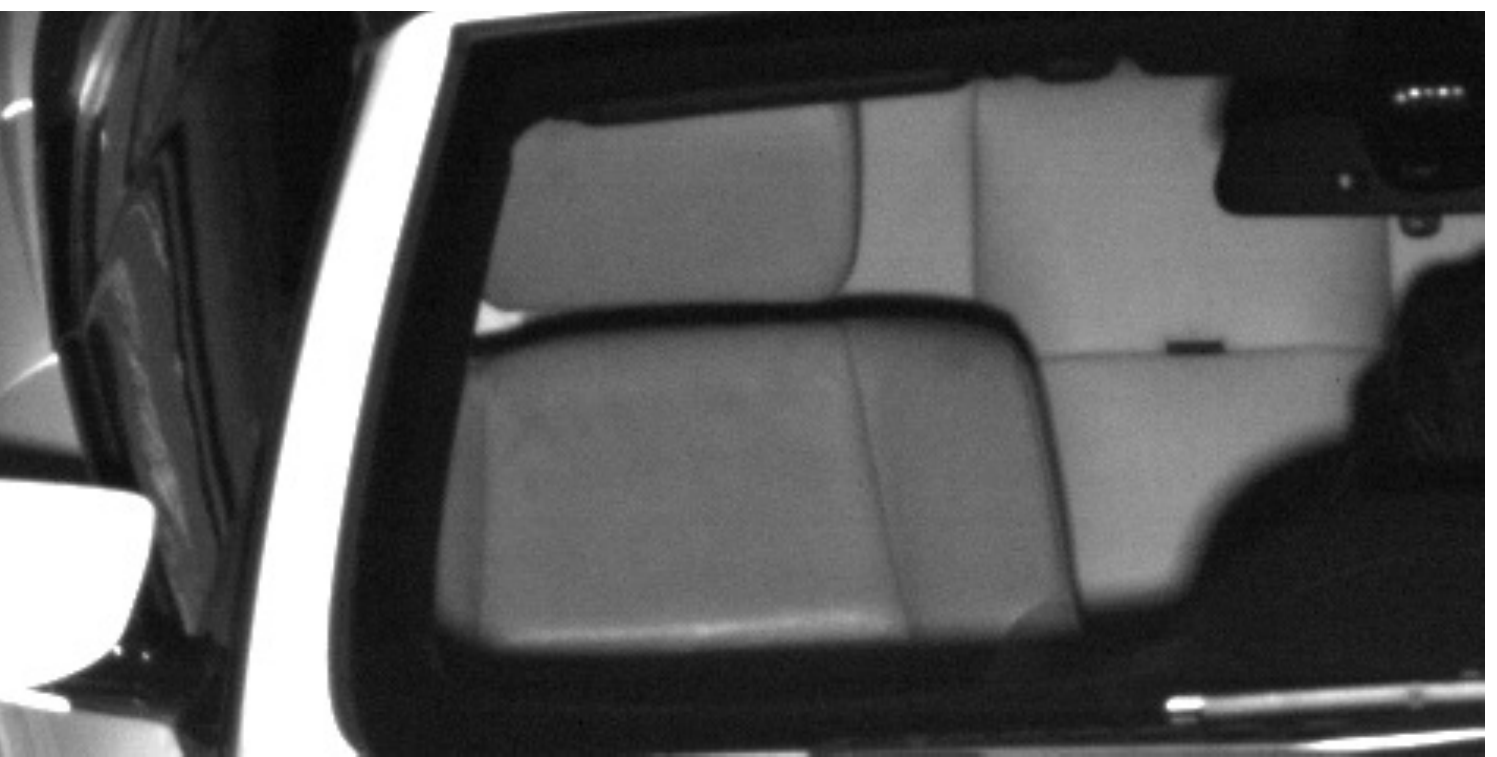}  & Empty & Empty \\ 
\hline
\includegraphics[width=0.8in, height=0.6in]{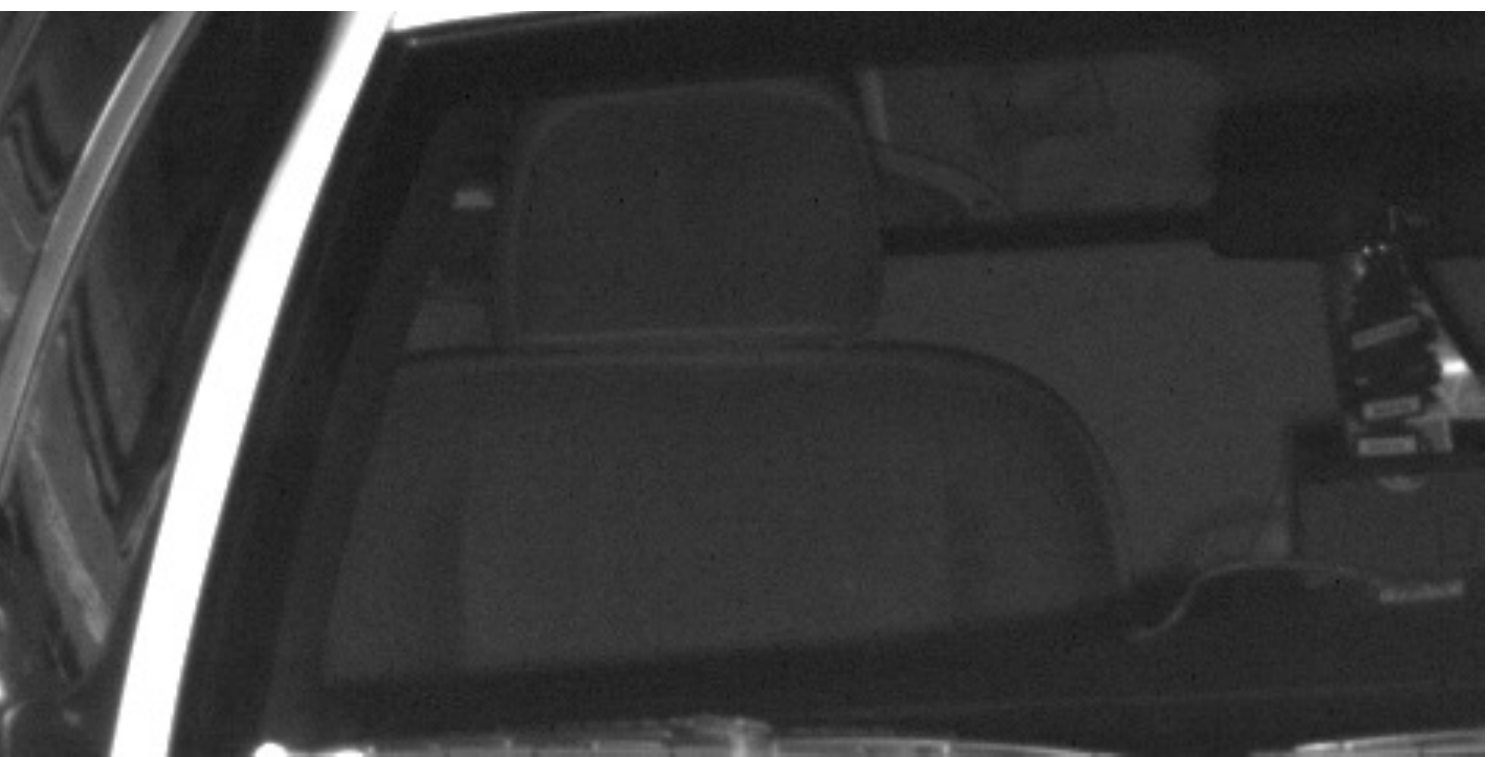} & Empty & Empty \\ 
\hline
\includegraphics[width=0.8in, height=0.6in]{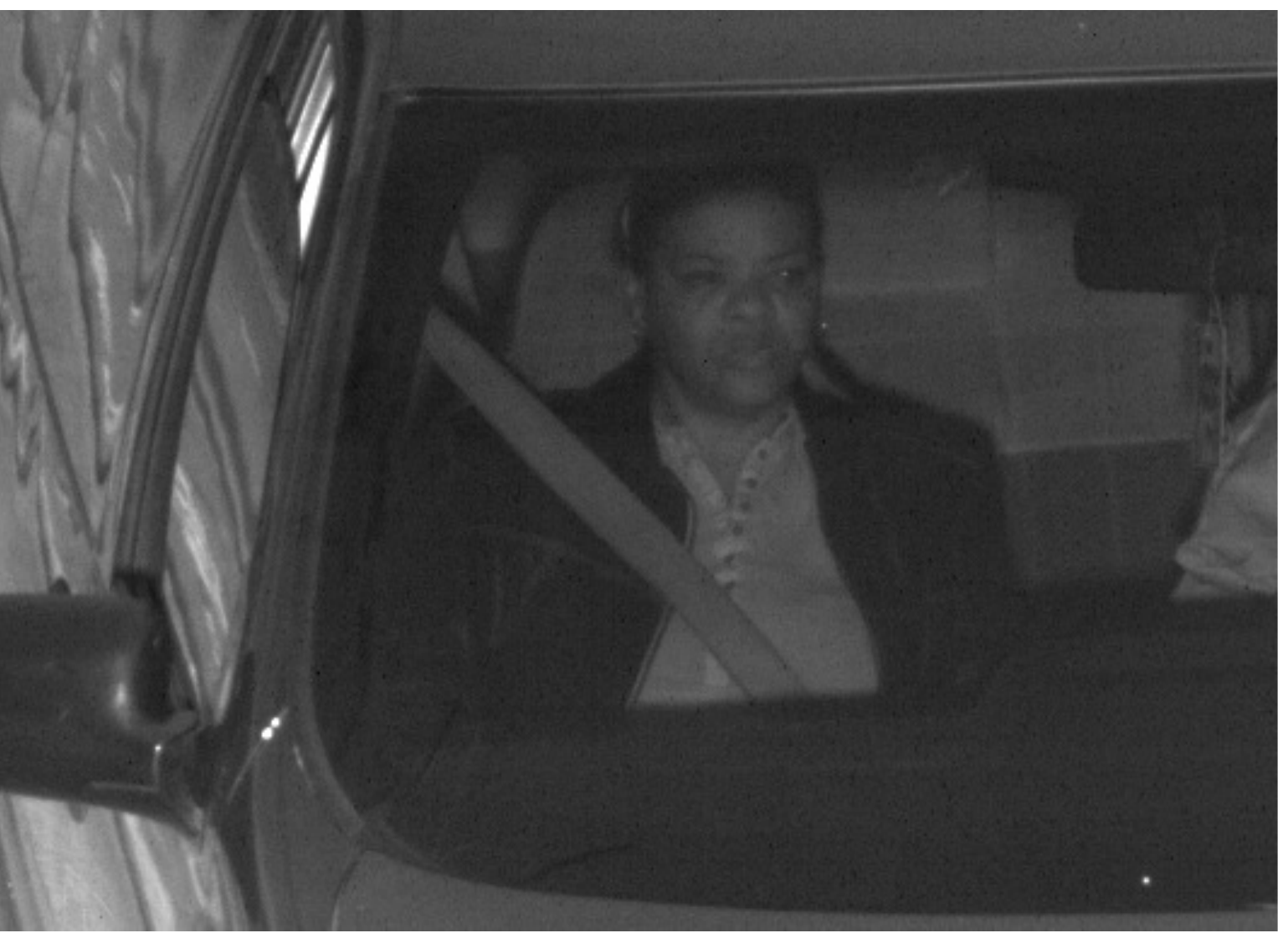}  & Person & Person \\ 
\hline
\includegraphics[width=0.8in, height=0.6in]{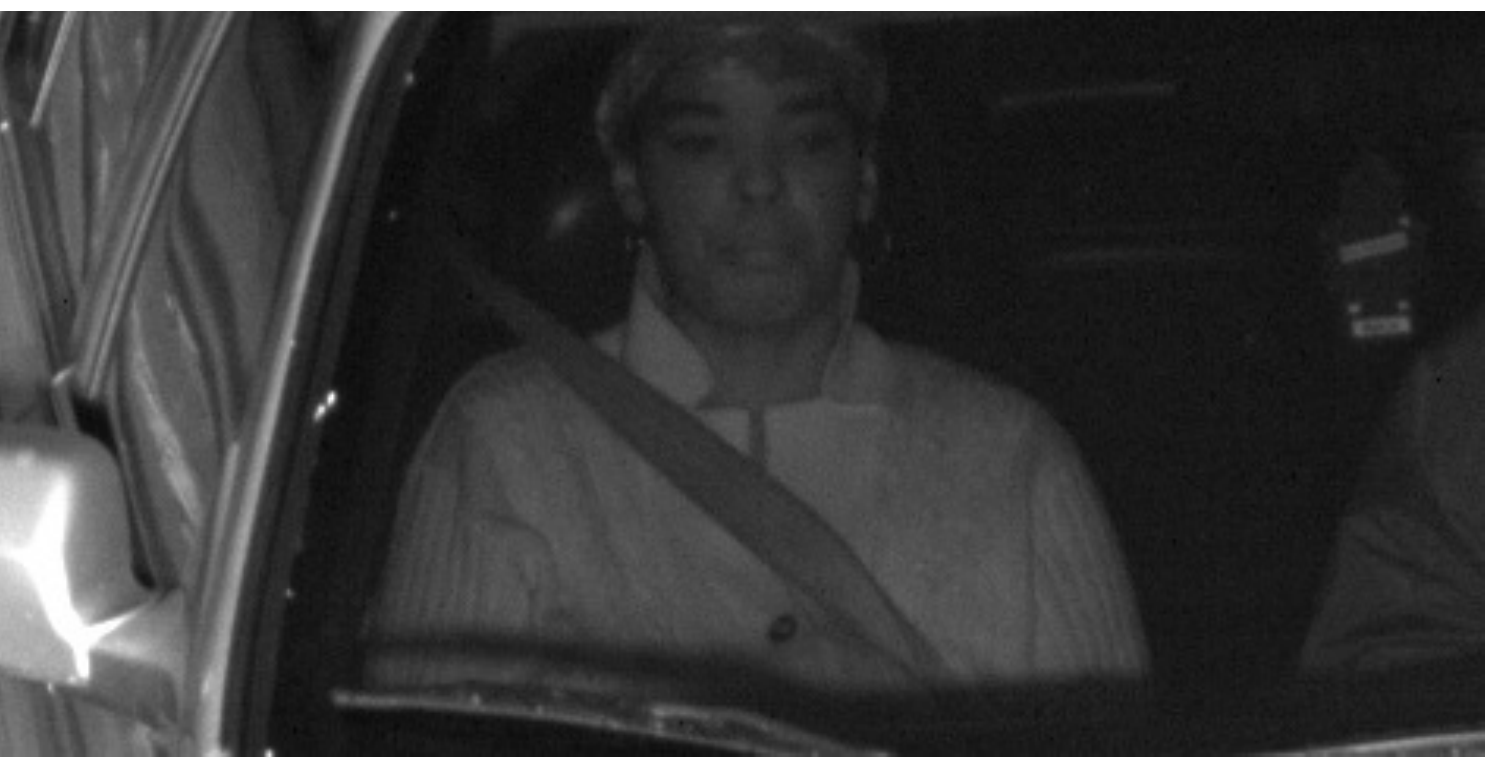}  & Person & Person \\ 
\hline
\end{tabular}
}
\caption{ \textbf{ For a given image, we present the face detection \cite{Zhu} versus image classification using FV results. Best threshold is chosen for face detection method. Images in Column 1 are contrast enhanced to show the passenger.}
}
 \label{fig:TestFigureAll}
\end{figure*}

VLAD is a simplified non probabilistic version of the Fisher Vector (FV). Similar to BoW, a codebook is generated using k-means clustering \cite{Perronin3}. Each local descriptor $x_{t}$ is associated to its nearest visual word in the codebook. For each visual codeword $\mu_{i}$, the differences $x_{t}-\mu_{i}$ of the vectors $x_{t}$ assigned to $\mu_{i}$ are accumulated to form a vector $v_{i}$ as shown in Eq. \ref{eq:vlad} 
\begin{equation}\label{eq:vlad}
v_{i} = \sum_{x_{t}:NN(x_{t})=i}x_{t}-\mu_{i}
\end{equation}
The VLAD is formed by the concatenation of the $d$-dimensional vectors $v_{i}$. Similar to FV, we apply the power and L2 normalization on VLAD vectors. 
\newline
\indent  
Once the features are learned (for positive and negative training images) using one of the local aggregation methods presented above, we learn a classifier model utilizing a linear SGD-SVM with hinge loss using the stochastic gradient descent techniques \cite{Zeynep}, \cite{Bottou}.\newline
 \indent
Table \ref{tab:table_CompareMethods} presents classification accuracy using various local aggregation methods. Fisher vectors outperforms other methods including face detection by achieving an accuracy of approximately 96\% using $K=256$. Figure \ref{fig:VLADvsFV} shows that performance of FV is consistently better than VLAD if the $K$ value is set above 64. Similar to \cite{Perronin3}, we have also performed PCA on the FV and VLAD vectors to observe the effect of dimension reduction in the overall classification performance (we reduced number of dimensions from $K\times d$ to 512 for both FV and VLAD), FV+PCA achieves an accuracy of 93.72 \% at K=256. Similarly, FV+PCA consistently performs better than VLAD+PCA for this study. \newline
\indent
Figure \ref{fig:ROCs} shows the receiver operating characteristics (ROC) curves and Accuracy versus Yield curves for FV, VLAD, BoW and face detection method of \cite{Zhu}. Note that face detector performs poorly compare to the image classification based methods. However, if we set the number of gaussians in FV too low (such as $K=32$), detection rate degrades significantly. Note also that at the yield level of 80\%, we can achieve over 99\% accuracy using the image classification based on FV and VLAD representations. 
Figure \ref{fig:TestFigureAll} presents several samples of the occupancy detection task in HOV/HOT lanes. FV has correctly determined the presence of passengers in Column $1$ images while face detection method of \cite{Zhu} failed. Rows $3$, $4$ and $5$ (of Column $1$) shows the robustness of this approach even in the presence of the occlusion.
 
\begin{table*}[!t]
\centering
   \caption{Classification accuracies of Bag-of-words (BoW), Fisher vectors (FV), and VLAD for various number of gaussians (K).}
    \begin{tabular}{|c|c|c|c|c|c|c|c|c|}
    \hline
     $K$ &  32 & 64 & 128 & 256 & 512 & 1024 & 2048 & 4096  \\
  \hline		
     $\mathrm{BoW}$ &  0.9376  & 0.9412 & 0.9384 & 0.9348 & 0.9464 & 0.9504 & 0.9528 & 0.9424  \\
 \hline
     $\mathrm{VLAD}$ &  0.9244 & 0.9312 & 0.9444 & 0.9496 &  0.9508 & 0.9372 & - & - \\
 \hline   
     $\mathrm{Fisher}$ & 0.7513 & 0.9460 & 0.9540  & 0.9596 & 0.9544 & - & - & -  \\
 \hline
    \end{tabular}
\label{tab:table_CompareMethods}
\end{table*}

\epsfxsize 0.8in \epsfysize 0.6in

\section{Discussion \& Conclusion} \label{sec:Discussion}
There are several key advantages of the image based classification approach compared to the face detection methods. First, it should be noted that the face does not need to be located within the scene. Face location is a time consuming and computationally costly process that is inherent to face detection methods. Second, non-facial passenger information such as torsos, arms, shoulders, etc. naturally are used in the image classification approach, while not taken advantage of in the face detection approach. In order to incorporate non-face related structure to the deformable part model based face detection methods, model complexity has to be increased significantly, with the related ground truthing complexity involved.\newline
\indent 
In this paper, we compare face detection and image classification for use in the front seat vehicle occupancy detection task. Experiments using 3000 real-world images captured on a city roadway indicate that the image classification approach is far superior for front seat occupancy detection. The state-of-the-art deformable part model based approach for face detection \cite{Zhu} underperforms a Fisher vector based image classification approach on this dataset. In further testing across differing roadways for tens of thousands of images, the FV approach has continued to yield accuracy rates above 95\% for the front seat vehicle occupancy task.


\begin{thebibliography}{}\itemsep=-1pt

\bibitem{Xerox} http://services.xerox.com/transportation-solutions/transportation-management/enus.html

\bibitem{Stephen}S.~Schijns, P.~Mathews, ``A Breakthrough in Automated Vehicle Occupancy Monitoring Systems for HOV / HOT Facilities", \textit{12th HOV Systems Conference}, April 20, 2005.

\bibitem{Billheimer}J.~Billheimer, K.~Kaylor, C.~Shade, ``Use of videotape in HOV lane surveillance and enforcement," \textit{Technical Report, U.S. Department of Transportation}, 1990.

\bibitem{Zhu} X. Zhu, D.~Ramanan, â``Face Detection, Pose Estimation, and Landmark Localization in the Wild," in \textit{CVPR}, 2012.

\bibitem{Csurka} G.~Csurka, C.~Dance, L.~Fan, J.~Willamowski, C.~Bray, ``Visual Categorization with bag of keypoints," in \textit{ECCV SLCV Workshop}, 2004. 

\bibitem{Jaakkola}T.~Jaakkola, D.~Haussler, ``Exploiting generative models
in discriminative classifiers," in \textit{NIPS}, Dec. 1998.

\bibitem{Perronin1}F.~Perronnin, C.~R.~Dance, ``Fisher kernels on visual vocabularies
for image categorization," in \textit{CVPR}, June 2007

\bibitem{Perronin2}F.~Perronnin, J.~Sanchez,  T.~Mensink, ``Improving Fisher Kernels for Large-Scale Image Classification," in \textit{ECCV},
2010.

\bibitem{Perronin3}H.~Jegou, F.~Perronin, M.~Douze, J.~Sanchez, P.~Perez, C.~Schmid,``Aggregating local image descriptors into compact codes," \textit{IEEE PAMI}, 2011.

\bibitem{Felzenszwalb}P.~F.~Felzenszwalb, R.~B.~Grishick, D.~McAllester, D.~Ramanan, ``Object detection with discriminatively trained part
based models," \textit{IEEE PAMI} , 2009.

\bibitem{Dalal} N.~Dalal, B.~Triggs, ``Histograms of oriented gradients for human detection," in \textit{CVPR}, 2005.

\bibitem{Jones}M.~Jones, P.~Viola, ``Fast multi-view face detection," in \textit{CVPR}, 2003. 

\bibitem{Jegou}H.~Jegou, M.~Douze, C.~Schmid, ``Improving bag-of-features for large scale image search," \textit{International Journal of Computer Vision}, vol. 87, pp. 316-336, Feb. 2010.

\bibitem{SIFT}D.~Lowe, ``Distinctive image features from scale invariant from scale-invariant keypoints,"  \textit{International Journal of Computer Vision}, vol. 60, no. 2, pp. 91-110, 2004.

\bibitem{Sivic}J.~Sivic, A.~Zisserman, ``Video Google: A text retrieval approach to object matching in videos," in \textit{ICCV}, pp. 1470-1477, Oct. 2003. 

\bibitem{Zeynep}F.~Perronnin, Z.~Akata, Z.~Harchaoui, C.~Schmid, ``Towards good practice in large-scale learning for image classification," in \textit{CVPR}, 2012.

\bibitem{Bottou}L.~Bottou, ``Stochastic Learning," \textit{Advanced Lectures on Machine Learning}, pp. 146-168, 2003.

\bibitem{Peter1} E.~N.~Dalal., P.~Paul, L.~K.~Mestha, A.~S.~Islam, ``Vehicle OccupancyDetection via single band infrared imaging," United States Patent Application, 20130141574, 2013.
 
\bibitem{Peter2} Z.~Fan, A.~S.~Islam, P.~Paul, B.~Xu, L.~K.~Mestha, ``Front seat vehicle occupancy detection via seat pattern recognition,"  United States Patent Application, 20130051625, 2013.

\bibitem{Hao}X.~Hao, H.~Chen, J.~Li, ``An Automatic Vehicle Occupant Counting Algorithm Based on Face Detection," in \textit{Int. Conf. on Sign. Proc.}, vol 3, 2006. 

\bibitem{Birch}P.~M.~Birch, R.~C.~D.~Young, F.~Claret-Tournier, C.~R.~Chatwin, ``Automated Vehicle Occupancy Monitoring," \textit{Optical Engineering}, vol 43, no. 8, pp. 1828-1832, 2004.

\bibitem{Georgia} W.~Daley, O.~Arif, J.~Stewart, J.~Wood, C.~Usher, E.~Hanson, J.~Turgeson, D.~Britton, ``Sensiting system development for HOV (High Occupancy Vehicle) Lane Monitoring," \textit{Georgia Department of Transportation Technical Report}, 2011.

\end{thebibliography}
\end{document}